\documentclass[conference]{IEEEtran}
\pdfoutput=1
\IEEEoverridecommandlockouts
\usepackage{cite}
\usepackage{amsmath,amssymb,amsfonts}
\usepackage{algorithm}
\usepackage{algpseudocode}
\usepackage{graphicx}
\usepackage{textcomp}
\usepackage{xcolor}
\def\BibTeX{{\rm B\kern-.05em{\sc i\kern-.025em b}\kern-.08em
    T\kern-.1667em\lower.7ex\hbox{E}\kern-.125emX}}
\usepackage{cuted}
\usepackage{capt-of}
\usepackage{balance}
\usepackage[letterpaper]{geometry}
\usepackage{hyperref}
\usepackage{xcolor}

\begin{document}


\newcommand{\cmt}[1]{}
\newcommand{\ren}[1]{\textcolor{orange}{{Ren: #1}}}
\newcommand{\sehoon}[1]{\textcolor{red}{{Sehoon: #1}}}
\newcommand{\nitish}[1]{\textcolor{magenta}{{Nitish: #1}}}
\newcommand{\revised}[1]{\textcolor{black}{{#1}}}
\newcommand{\original}[1]{\textcolor{magenta}{Original: #1}}
\newcommand{\eqnref}[1]{Equation~(\ref{eq:#1})}
\newcommand{\figref}[1]{Figure~\ref{fig:#1}}
\newcommand{\tabref}[1]{Table~\ref{tab:#1}}
\newcommand{\secref}[1]{Section~\ref{sec:#1}}

\long\def\ignorethis#1{}

\newcommand{\etal}{{\em{et~al.}\ }}
\newcommand{\eg}{e.g.\ }
\newcommand{\ie}{i.e.\ }

\newcommand{\figtodo}[1]{\framebox[0.8\columnwidth]{\rule{0pt}{1in}#1}}



\newcommand{\pdd}[3]{\ensuremath{\frac{\partial^2{#1}}{\partial{#2}\,\partial{#3}}}}

\newcommand{\mat}[1]{\ensuremath{\mathbf{#1}}}
\newcommand{\set}[1]{\ensuremath{\mathcal{#1}}}

\newcommand{\vc}[1]{\ensuremath{\mathbf{#1}}}
\newcommand{\vEndEff}{\ensuremath{\vc{d}}}
\newcommand{\vRelMove}{\ensuremath{\vc{r}}}
\newcommand{\sSet}{\ensuremath{S}}

\newcommand{\vControl}{\ensuremath{\vc{u}}}
\newcommand{\vPoint}{\ensuremath{\vc{p}}}
\newcommand{\sSpringCoef}{{\ensuremath{k_{s}}}}
\newcommand{\sDamperCoef}{{\ensuremath{k_{d}}}}
\newcommand{\vHandle}{\ensuremath{\vc{h}}}
\newcommand{\vForce}{\ensuremath{\vc{f}}}

\newcommand{\mTransChain}{\ensuremath{\vc{W}}}
\newcommand{\mRotateTrans}{\ensuremath{\vc{R}}}
\newcommand{\sJoint}{\ensuremath{q}}
\newcommand{\vJoint}{\ensuremath{\vc{q}}}
\newcommand{\mJoint}{\ensuremath{\vc{Q}}}
\newcommand{\mMass}{\ensuremath{\vc{M}}}
\newcommand{\sMass}{\ensuremath{{m}}}
\newcommand{\vGravity}{\ensuremath{\vc{g}}}
\newcommand{\vConstr}{\ensuremath{\vc{C}}}
\newcommand{\sConstr}{\ensuremath{C}}
\newcommand{\vCOM}{\ensuremath{\vc{x}}}
\newcommand{\sGeneralForce}[1]{\ensuremath{Q_{#1}}}
\newcommand{\vStateVar}{\ensuremath{\vc{y}}}
\newcommand{\vControlVar}{\ensuremath{\vc{u}}}
\newcommand{\tr}[1]{\ensuremath{\mathrm{tr}\left(#1\right)}}

%
%

\renewcommand{\choose}[2]{\ensuremath{\left(\begin{array}{c} #1 \\ #2 \end{array} \right )}}

\newcommand{\gauss}[3]{\ensuremath{\mathcal{N}(#1 | #2 ; #3)}}

\newcommand{\pctab}{\hspace{0.2in}}
\newenvironment{pseudocode} {\begin{center} \begin{minipage}{\textwidth}
                             \normalsize \vspace{-2\baselineskip} \begin{tabbing}
                             \pctab \= \pctab \= \pctab \= \pctab \=
                             \pctab \= \pctab \= \pctab \= \pctab \= \\}
                            {\end{tabbing} \vspace{-2\baselineskip}
                             \end{minipage} \end{center}}
\newenvironment{items}      {\begin{list}{$\bullet$}
                              {\setlength{\partopsep}{\parskip}
                                \setlength{\parsep}{\parskip}
                                \setlength{\topsep}{0pt}
                                \setlength{\itemsep}{0pt}
                                \settowidth{\labelwidth}{$\bullet$}
                                \setlength{\labelsep}{1ex}
                                \setlength{\leftmargin}{\labelwidth}
                                \addtolength{\leftmargin}{\labelsep}
                                }
                              }
                            {\end{list}}
\newcommand{\newfun}[3]{\noindent\vspace{0pt}\fbox{\begin{minipage}{3.3truein}\vspace{#1}~ {#3}~\vspace{12pt}\end{minipage}}\vspace{#2}}

\newcommand{\key}{\textbf}
\newcommand{\fun}{\textsc}



\title{
\vspace{18pt}\LARGE \bf PM-FSM: Policies Modulating Finite State Machine \\ for Robust Quadrupedal Locomotion
}

\author{
Ren Liu,
Nitish Sontakke,
Sehoon Ha
\thanks{Georgia Institute of Technology, Atlanta, GA, 30308, USA}
\thanks{{\tt\small rliu384@gatech.edu, nitishsontakke@gatech.edu, sehoonha@gatech.edu}}
}


\maketitle

\begin{strip}\centering
\vspace{-40pt}
\includegraphics[width=\linewidth]{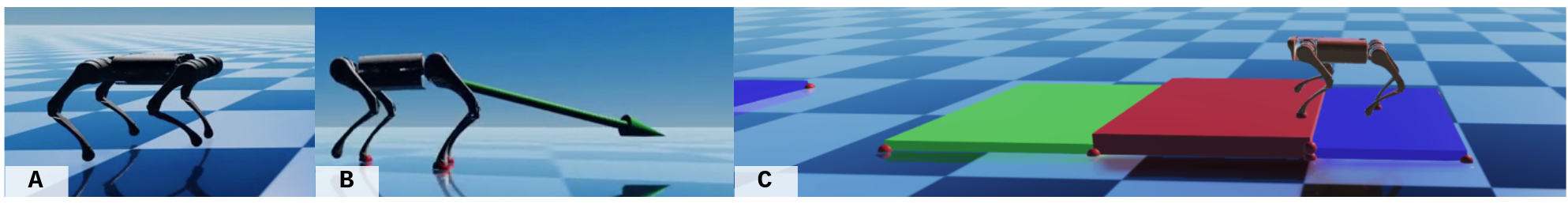}
\captionof{figure}{
Locomotion tasks covered in our simulated experiments, including the flat terrain (A), the flat terrain with external perturbations (B) and the randomized upstairs/downstairs terrains (C). Our proposed PM-FSM model proves its robustness on these simulation tasks, as well as real-world experiments.}
\label{fig:teaser}
\end{strip}

\begin{abstract}
Deep reinforcement learning (deep RL) has emerged as an effective tool for developing controllers for legged robots. However, vanilla deep RL often requires a tremendous amount of training samples and is not feasible for achieving robust behaviors. Instead, researchers have investigated a novel policy architecture by incorporating human experts' knowledge, such as Policies Modulating Trajectory Generators (PMTG). This architecture builds a recurrent control loop by combining a parametric trajectory generator (TG) and a feedback policy network to achieve more robust behaviors.
In this work, we propose Policies Modulating Finite State Machine (PM-FSM) by replacing TGs with contact-aware finite state machines (FSM), which offers more flexible control of each leg.
This invention offers an explicit notion of contact events to the policy to negotiate unexpected perturbations. We demonstrated that the proposed architecture could achieve more robust behaviors in various scenarios, such as challenging terrains or external perturbations, on both simulated and real robots. 

\end{abstract}


\section{Introduction}

Locomotion has been one of the most challenging problems in the robotics domain. A controller must achieve its primary task of moving its body toward the desired direction while maintaining the balance with limited sensing and actuation capabilities. 
One popular approach is model-based control that leverages identified dynamics and control principles, demonstrating effective locomotion on quadrupedal \cite{hutter2014quadrupedal, park2017high} and bipedal robots \cite{ramezani2014performance, westervelt2018feedback}. However, model-based control often requires considerable manual effort to develop a proper model for each task. On the other hand, deep reinforcement learning (deep RL) has emerged as a promising approach to learn a robust policy by maximizing the reward \cite{tan2018simtoreal, haarnoja2018learning, Hwangbo_2019}. To cope with extrapolated tasks, researchers have investigated more and more complex policies, such as long term short memory~\cite{hochreiter1997long} or temporal convolutional neural networks~\cite{lea2017temporal}. These policies require even more training samples and computations resources to obtain better performance, compared with common feed-forward network architectures.

Researchers have attempted to develop a hybrid technique to take the best from both model-based and learning-based approaches to reach both performance and time efficiency. For instance, a few prior works proposed hierarchical controllers where a policy outputs high-level parameters to low-level model-based controllers that are typically implemented with the Model Predictive Control strategy (MPC) \cite{li2017deepgait,tsounis2020deepgait,xie2021glide}. Although this approach is known to be sample-efficient and robust, its performance highly depends on the design of the low-level controller. Alternatively, researchers have investigated  Policies Modulating Trajectory Generators (PMTG)~\cite{iscen2019policies,lee2020learning,yang2020data} that embeds prior knowledge of trajectory generators into neural network policies and demonstrated great sample efficiencies.

Our key intuition is that the explicit notion of contacts can greatly improve the robustness of control policies \cite{bledt2018contact, park2012finite}, just like other model-based controllers. In PMTG, the predefined trajectory generator offers both memory and prior knowledge to the control policy to handle periodic locomotion. Still, it is difficult to learn very discrete behaviors due to the smooth nature of neural networks, even if we provide contact flags as additional inputs. On the other hand, finite state machines (FSM) can be more expressive by modeling abrupt behavior changes conditioned on contact signals.




In this work, we propose a novel policy representation that is explicitly aware of locomotion context including foot contact flags. We propose a novel policy architecture, Policies Modulating Finite State Machine (PM-FSM), which extends the trajectory generators in the original PMTG. Our key idea is to replace trajectory generators with finite state machines. This simple extension allows a robot to be aware of contact events explicitly and effectively adapt its behaviors to perturbed scenarios. 


We evaluate the proposed PM-FSM on the locomotion task of a quadrupedal robot, A1, in both simulated and real-world environments. Our results suggest that our architecture shows more robust behaviors than the original PMTG and its variant in various scenarios with external perturbations, and is better at the sim-to-real transfer. We then show that some complicated reflexes, such as going upstairs and downstairs, can also be learned using PM-FSM.

\section{Related Works}
\subsection{Locomotion with FSM}
It has been a very long time for researchers to use FSMs to make robots walk. Mcghee \etal \cite{tomovic1966finite,mcghee1967finite,mcghee1968some} are some of the earliest studies to prove that simple FSMs can accomplish the coordination of effective joint movements. Mcghee \etal\cite{mcghee1968some} defined each leg of the quadrupedal as a two-state automata. They also developed a hierarchical structure where the states of the hip and knee joints act as a sub-automata of the leg belonged to. 
Since then, more complicated controllers based on finite state machines have been developed for various applications related to robotic locomotion. Park \etal\cite{park2012finite} adopted the idea of sensory reflex-based control strategy \cite{huang2005sensory} and designed an finite state machine to manage unexpected large ground-height variations in bipedal robot walking. And Lee \etal\cite{lee2010data} showed that finite state machines can also be used in a data-driven biped control to generate robust locomotion. Recently, Bledt \etal\cite{bledt2018contact} showed that locomotion context like contact states of each leg can be used in finite state machines to help to generate stable locomotion and to assist gait switching. However, the design of effective FSMs requires a lot of time-consuming trials and errors based on prior knowledge if the goal is to overcome challenging terrains or dynamic environments.
\subsection{Locomotion with Deep RL}
The recent advances of deep reinforcement learning enable a more convenient approach for developing control policies by leveraging simple reward descriptions. Various policy gradient methods, such as Deep Deterministic Policy Gradient  (DDPG)~\cite{lillicrap2019continuous}, Trust Region Policy Optimization (TRPO)~\cite{schulman2017trust} and Proximal Policy Optimization (PPO)~\cite{schulman2017proximal}, have been widely adopted to train effective control policies, particularly in the context of robotic locomotion. As these methods usually learn policies in simulated environments, a lot of efforts to overcome the \textit{reality gap} have been made, such as domain randomization \cite{tan2018simtoreal,chebotar2019closing,xie2020dynamics}, meta learning \cite{finn2017modelagnostic,yu2020learning}, and compact observation space \cite{tan2018simtoreal}. Recently, a few learning-based approaches including training an extra network to model actuator dynamics~\cite{Hwangbo_2019} or training a Cycle-GAN that maps the images from the simulator to corresponding realistic images \cite{rao2020rlcyclegan} have also reached great performance in real robot experiments. 

Although simple MLPs have been used as policy network in a large number of studies \cite{tan2018simtoreal,chebotar2019closing,xie2020dynamics}, PMTG\cite{iscen2019policies} offers an effective approach to improve the performance of this simple policy architecture by taking an advantage of predefined trajectory generators (TGs). In this architecture, the policy can directly modulate the behavior of the predefined trajectory generators while generating additional feedback control signals. It has been shown that PMTG architecture can produce robust locomotion policies in both simulated and real-world environments.

\vspace{0.5em}
In this paper, we propose a new policy architecture, PM-FSM, by combining both directions, FSMs and deep RL. The prior knowledge is presented as a contact-aware FSM, of which the state transition functions are defined according to the robot proprioceptive sensory information (foot contact flags and target joint angles). This FSM is modulated with a feedback network which learns a robust control policy using deep RL. 

\section{Policies Modulating Finite State Machine}


\subsection{Background: PMTG}



PMTG~\cite{iscen2019policies} divides policy outputs into the trajectory generator modulating parameters ($\rho$) and the feedback terms ($u_{fb}$). The policy modulating parameters $\rho$ include the trajectory generator (TG)'s frequency $f$, amplitude $A$, and height $h$. Then the TG converts the parameters into control signals $u_{tg}$ and add them with the feedback terms. This formulation allows a policy to explore smooth behaviors by leveraging the pre-defined TG, while fine-tuning the behaviors with the feedback terms. Typically, PMTG is optimized with on-policy RL algorithms, such as PPO~\cite{schulman2017proximal} and ARS~\cite{mania2018simple}. For more details, please refer to the original paper.


\begin{figure}[ht]
    \centering
    \includegraphics[width=\linewidth]{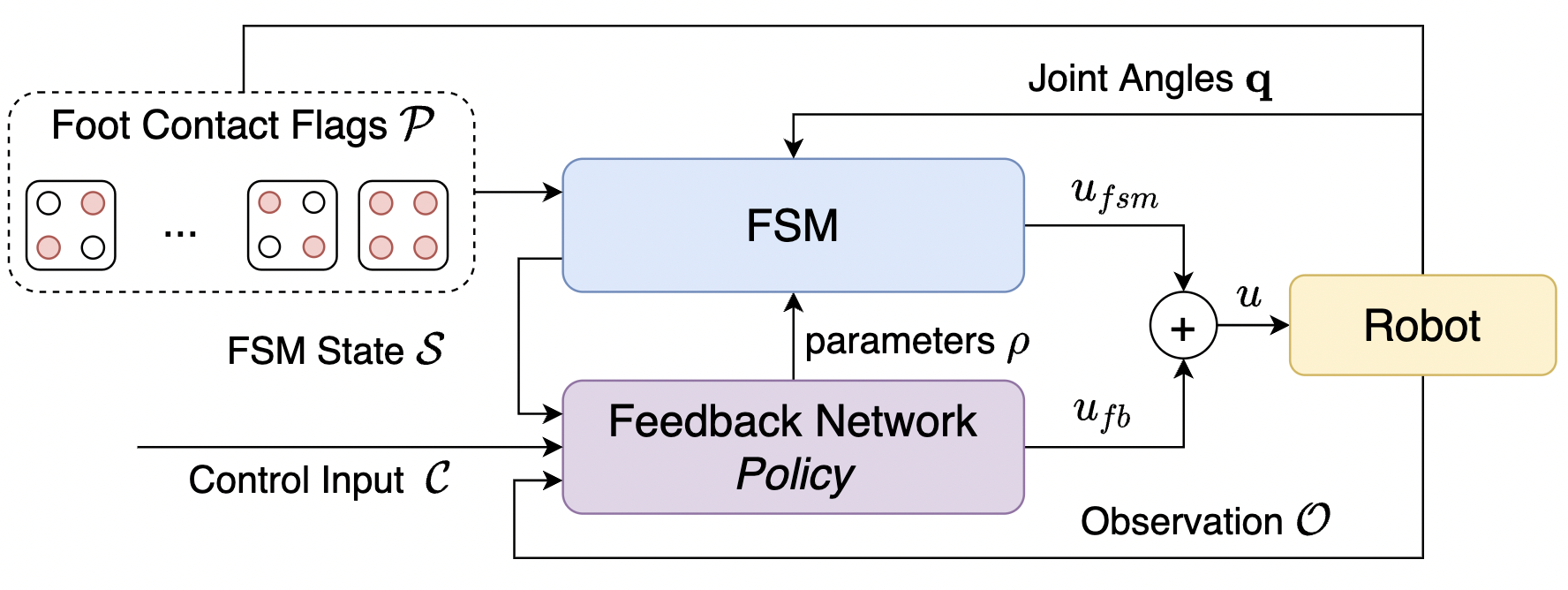}
    \caption{Overview of PM-FSM: The output (actions) $u_{fsm}$ of a predefined FSM combined with that of a learned feedback network ($u_{fb}$). The learned policy also modulates the parameters $\rho$ of the FSM at each time step and observes its state $\mathcal{S}$. The state transition conditions of the FSM are based on the current contact flags $\mathcal{P}$ and joint angles  $\vc{q}$.}
    \label{fig:pmfsm}
\end{figure}
\begin{figure*}[!ht]
    \centering
    \includegraphics[width=0.98\textwidth]{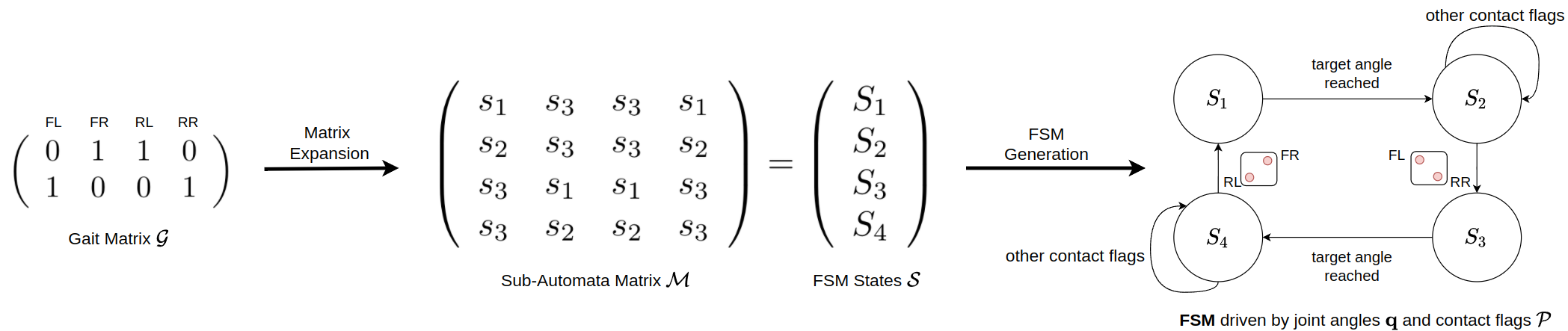}
    \caption{Our Finite State Machine Design. We expand the given gait matrix $\mathcal{G}$ by dividing the transfer phase ($0$) to  leg extension ($s_1$) and leg retraction ($s_2$) states. We keep the support phase ($1$) as leg angle adjustment state ($s_3$). Then every row in the sub-automata matrix $\mathcal{M}$ forms an FSM state. FL: Front Left, FR: Front Right, RL: Rear Left, RR: Rear Right. }
    \label{fig:my_fsm}
\end{figure*}
\subsection{Overview of PM-FSM}
Our key idea is to further extend the TG of the PMTG with an FSM that is explicitly aware of contacts. The top-level architecture for PM-FSM is shown in Figure \ref{fig:pmfsm}. Compared with  PMTG, our FSM explicitly considers locomotion contexts $\Sigma$ that consists of the current foot contact flags $\mathcal{P}$ and joint angles $\vc{q}$. Note that these locomotion contexts are directly read from the robot sensors and does not need to be part of the policy inputs, although they often overlap in practice.
    
\subsection{Policy Modulating Parameters}
In our system, the policy takes FSM state $\mathcal{S}$ and generates the policy modulating term $\rho=(f, A, h)$, frequency $f$, amplitude $A$, and height, along with the feedback term $u_{fb}$). Then FSM combines these parameters and locomotion contexts $\Sigma = (\mathcal{P}, \vc{q})$ to configure each leg motion generator and predict next actions $u_{fsm}$. Note that there are no learnable parameters within the FSM, which is similar to PMTG. FSM changes its behavior only via policy modulation.

In PMTG, the frequency $f_{tg}$ will influence the speed of joint rotation. As FSM controller does not have a shared system cycle time, we use $f$ to determine the ideal cycle step $T$ when the robot walks on the flat terrain without external perturbations. With control time interval as $dt$,  we define  the ideal cycle step $T$ as:
\begin{equation}
    T = \lceil \frac{1}{f \cdot dt} \rceil.
\end{equation} 
The amplitude $A$ and height $h$ will decide the target joint angles, which does not require a well-tuned configuration. 
   


\subsection{Design of Finite State Machine}

In this section, we will describe the gait-based FSM (Figure \ref{fig:my_fsm}). Our design is inspired by a gait matrix $\mathcal{G}$ introduced by Mcghee \etal~\cite{mcghee1968some}.
In their definition, each leg of the robot can be designed as a 2-state automata in locomotion. These two states are usually referred to as the \textit{support phase} and \textit{transfer phase}, which represent the state that supporting on the ground and swinging forehead separately. And a gait matrix $\mathcal{G}$ with \textit{k} columns is introduced to describe the locomotion automation composed of \textit{k} legs, where $\mathcal{G}_{ij}$ means in the gait state $i$, whether the leg $j$ is in support ($\mathcal{G}_{ij}=0$) or transfer ($\mathcal{G}_{ij}=1$) phase. 

\begin{figure}[ht]
    \centering
    \includegraphics[width=0.48\textwidth]{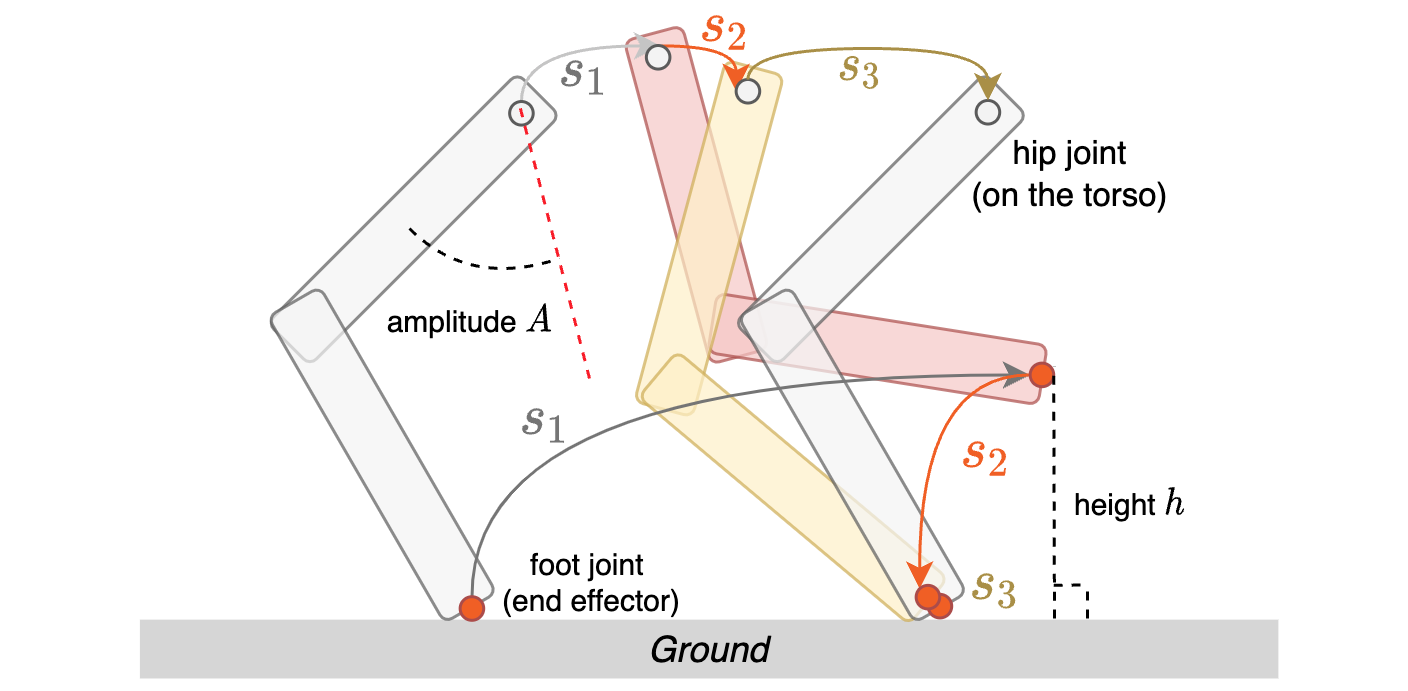}
    \caption{An Example of Joint Sub-Automata: \textbf{Leg extension[$s_1$]: }The leg moves from the most rear position to the most front position. \textbf{Leg retraction[$s_2$]: }The raised leg will move back to contact with the ground. \textbf{Leg angle movements[$s_3$]: } The foot joint will be kept on the ground while moving the hip/knee angles to lift and push forward the torso. The amplitude $A$ defines the targe hip angle difference between $s_1$ and $s_2$, and the height $h$ defines the distance from the lifted foot to the ground at the end of $s1$.}
    \label{fig:jointSubAutomata}
    \vspace{-1em}
\end{figure}

We further build a sub-automata matrix $\mathcal{M}$ by expanding the given gait matrix $\mathcal{G}$ by adopting a three-state joint sub-automata with the Swing Retraction model~\cite{seyfarth2003swing,wisse2005swing}. 
In this purpose, we divide the transfer phase ($0$) into leg extension ($s_1$) and a leg retraction ($s_2$) sub-automata states.
Note that this method requires that there is no successive transfer phases in any column of the gait matrix $\mathcal{G}$.
Then we can compute joint angles for each sub-automata states $s_1$, $s_2$, and $s_3$ using the FSM parameters $(f, A, h)$ given from a policy: please refer to Figure~\ref{fig:jointSubAutomata} for more details.
This expansion allows us a finer control over each leg.


Then we build the final FSM states $\mathcal{S}_i$ from each row of the expanded automata $\mathcal{M}$.
The start state of the FSM is set to $\mathcal{S}_1$ in default. 
The set of final states $\mathcal{F}=\phi$ because we aim to develop an indefinite walking controller.
Finally, we define the state transition functions as:
$$
\delta(\mathcal{S}_i,\mathcal{S}_{i+1})=
\begin{cases}
\text{all swing legs make contacts, if } \mathcal{S}_2 \text{ or } \mathcal{S}_4 \\
\text{all joints reach target angles, if }  \mathcal{S}_1 \text{ or } \mathcal{S}_3
\end{cases}
$$
Then this formulation defines the FSM as quintuple $(\Sigma, \mathcal{S}_1, \mathcal{S}, \delta, \mathcal{F})$.

\subsection{Reflexes}
The explicit notion of contact flags also allows us to easily extend PM-FSM with reflex controllers. Park \etal~\cite{park2012finite} proposed the idea of tripping reflex as part of the FSM to cope with complex tasks like going upstairs and downstairs. These model-based methods often require precise tuning to maximize their performance. However,  in our framework, we expect a feedback policy is capable of learning how to leverage heuristic reflex controllers.

\begin{figure}
    \centering
    \vspace{-0.5em}
    \includegraphics[width=0.48\textwidth]{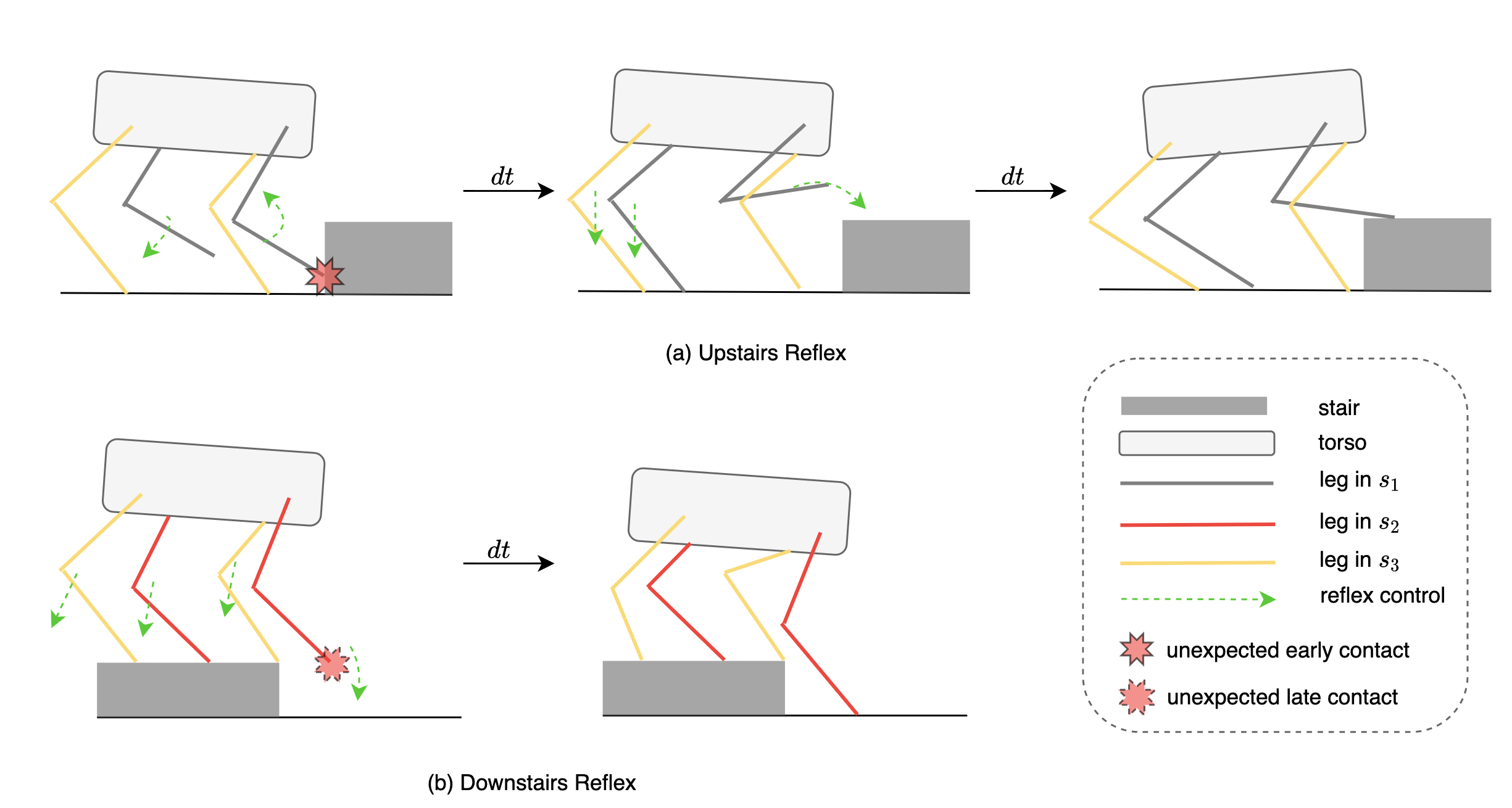}
    \caption{Upstairs reflex (UR) and downstairs reflex (DR) in gait-based FSMs: (a) UR: When unexpected early contact happens while the leg is in the leg extension state ($s_1$), first raise the collided leg higher and retract other extended legs, then bend the rear legs (RL, RR) and retry leg extension. (b) DR: When unexpected late contact happens while the leg is in the leg retraction state ($s_2$), extend the dangling leg and lower the torso.
    }
    \label{fig:reflex}
    \vspace{-1em}
\end{figure}

There are two simple reflexes designed in our FSM: going upstairs and downstairs reflexes. To activate the upstairs reflex (UR), we detect if there is an unexpected contact signal before the leg reaches its target joint angle during the leg extension state ($s_1$). To activate the downstairs reflex (DR), we detect whether the leg is dangling when the leg reaches its target joint angle at the end of the leg retraction state ($s_2$). In each case, we propose two-step and one-step reflex mechanisms for UR and DR, respectively. See Figure \ref{fig:reflex} for details.

Note that all parameters in these reflexes, including how much the collided leg will be lifted and how much the rear legs will bend, are determined heuristically without fine-tuning. These parameters are expected to be further adjusted by the feedback network. To make the policy be aware of applied reflexes, an additional \textit{reflex state} showing the current activated reflex's ID will be appended to FSM state $\mathcal{S}$ to be passed to the feedback network.

\begin{figure*}[!ht]
    \centering
    \includegraphics[width=0.97\textwidth]{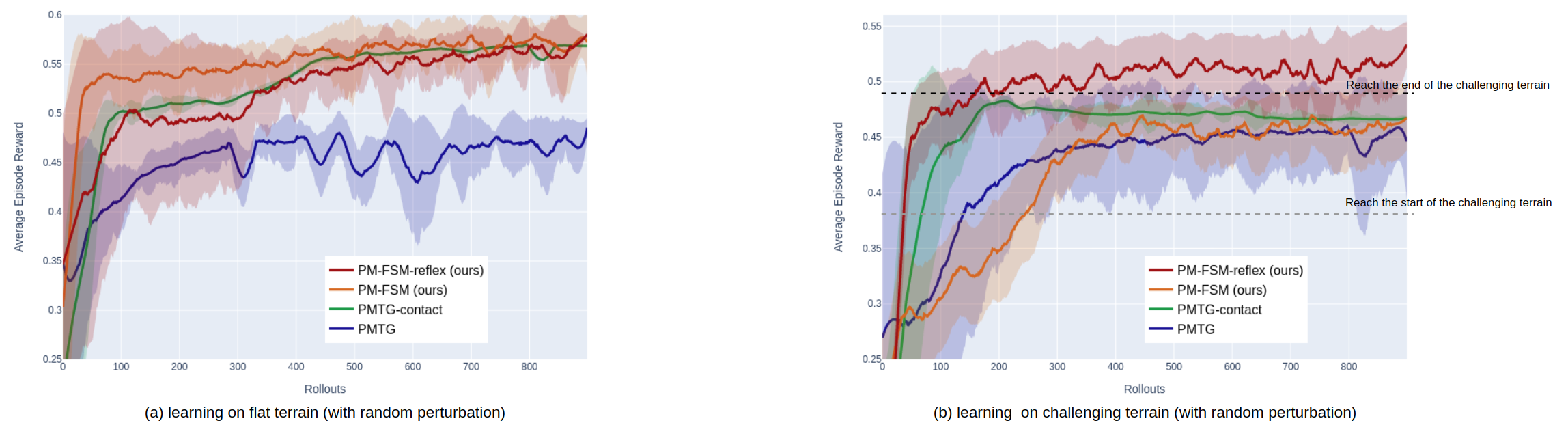}
    \caption{Learning curves: 
    (a) All PM-FSM-reflex (Ours), PM-FSM (Ours), and PMTG-contact learned a stable policy in the \emph{training} environment except for the vanilla PMTG without contact flags.
    (b) However, only PM-FSM-reflex agent learned an effective policy to go through the whole challenging scenarios with stairs and random perturbations. }
    \label{fig:trainingcurve}
    \vspace{-10px}
\end{figure*}
\subsection{Interpolated Motion Control}
In our design, we parameterize joint sub-automata with three target poses with six variables. When we want to control the robot, we do not want to use the given pose directly as PD target because it will cause unnecessarily abrupt movements. Instead, our goal is to apply exponential interpolation for achieving smoother motions by designing the following kinematic controller. For a $k$-state finite state machine, the duration distribution $\tau$ is defined by a $k$-dim vector
\begin{equation} \nonumber
    \tau = [d_1, ..., d_k], \,\, \text{where} \sum_{i=0}^{k} d_i = 1.
\end{equation}
and the termination position for each state is defined by the k-tuple of vectors $\epsilon=(\vc{e}_1, ..., \vc{e}_k)$. 

Let us assume that the current state is $i$, the current time is $t$ steps after entering the state $i$, and the current position is $\vc{p}(t)$, a simple exponential model is adopted to calculate the expected action $\vc{a}_t$, which suggests the increment from the current joint angles to the next angles:

\begin{equation} \label{eq:at}
    \vc{a}(t) = K_i \cdot (\vc{e}_i - \vc{p}(t)),
\end{equation}
where $K_i$ is a coefficient. We can solve $K_i$ to achieve the desired durations:

\begin{equation} \label{eq:KiFinal}
    K_i = 1 - \sqrt[\lfloor T \cdot d_i + 0.5 \rfloor]{\frac{\delta}{|e_i - e_{i-1}|}}
\end{equation}
\vspace{10px}

where $T$ is the total steps in the ideal situation and $\delta$ is the toleration. 

In our joint sub-automata, the duration distribution is set as a constant vector $\tau = [0.34, 0.16, 0.50]$, which is similar to the $1:2$ duration distribution for \textit{transfer phase} and \textit{support phase} mentioned in the work of Mcghee~\etal\cite{mcghee1967finite}.

\section{Experiments}
We conducted both simulated and real robot experiments to show that our PM-FSM allows us to learn robust quadrupedal locomotion policies on challenging terrains. Particularly, we designed our experiments to verify the following research questions:
\begin{enumerate}
    \item Can PM-FSM policies show more robust behaviors than PMTG policies?
    \item How do different design choices (like reflexes) affect the performance of PM-FSM?
\end{enumerate}

\subsection{Simulated Experiments}
\noindent \textbf{Experimental Setup.}
We took the RaiSim~\cite{raisim} as our training environment. The friction coefficient of the training terrains is set to $0.8$ to better simulate the blanket floor. 
We also added random directional perturbation forces, up to $10N$ horizontally and $30N$ vertically, to the robot multiple times at random timing to obtain more stable behaviors. To fit the given specification of AlienGo and A1 Explorer robots from Unitree \cite{Unitree}, we set the maximum velocity in the task $v_{max}$ as $0.6$ m/s. The control frequency is $40$ Hz and each rollout lasts for $25$ seconds. The total number of roll-out is set to $2000$.

We learned a policy with an Actor-Critic on-policy learning method, Proximal Policy Optimization. Both we represent the actor and critic networks as simple three-layer fully connected neural networks. The numbers of hidden nodes in two hidden layers are $128$ and $64$ respectively. We trained four policies for each experiment: 1)PMTG, 2)PMTG-contact, 3)PM-FSM, 4)PM-FSM-reflex. We did not compare them with a naive reactive agent without prior knowledge because they tend to require a far more simulation steps. All policies take $z$-axis of the robot frame in the world frame (3-dim), the current linear velocity of the robot along the target direction (1-dim), the current angular velocities of the robot (roll, pitch, yaw; 3-dim), and the target velocity generated by the velocity controller (1-dim) as input. Additionally, PMTG and PMTG-contact take a 2-dim embedding of TG phase~\cite{iscen2019policies}, and PMTG-contact takes extra 4-dim contact flags of each foot. Our PM-FSM and PM-FSM-Reflexes take the 4-dim sub-automata states for all legs, and PM-FSM-reflex takes an extra 1-dim reflex state as input FSM state. The output dimension is $11$ for all cases: $8$ dimensional $u_{fb}$ for pitch joint angles and $3$ dimensional $(f, A, h)$ for modulating parameters $\rho$.

Our reward function is designed to be sum of three terms: $r_{speed}$, $r_{torque}$ and $r_{done}$. The speed term $r_{speed}$ measures the difference between the current and target velocities, which is exactly the same as the reward defined in the original PMTG paper: 
\begin{equation}
    r_{speed} = v_{max} \exp(-\frac{(v_R-v_T)^2}{2v_{max}^2}),
\end{equation}
where $v_{max}$ is the maximum desired velocity for the task, and $v_R$ and  $v_T$ are the robot’s actual velocity and the target velocity at the current timestep. The torque term $r_{torque}$ is simply a negative coefficient $C_{\tau}$ times the squared sum of torques $\lVert \tau \rVert^2$. The termination penalty $r_{done}$ is $-0.5$ if the robot falls before the end of the rollout and is $0$ otherwise.

\begin{figure}[!ht]
    \centering
    \includegraphics[width=0.50\textwidth]{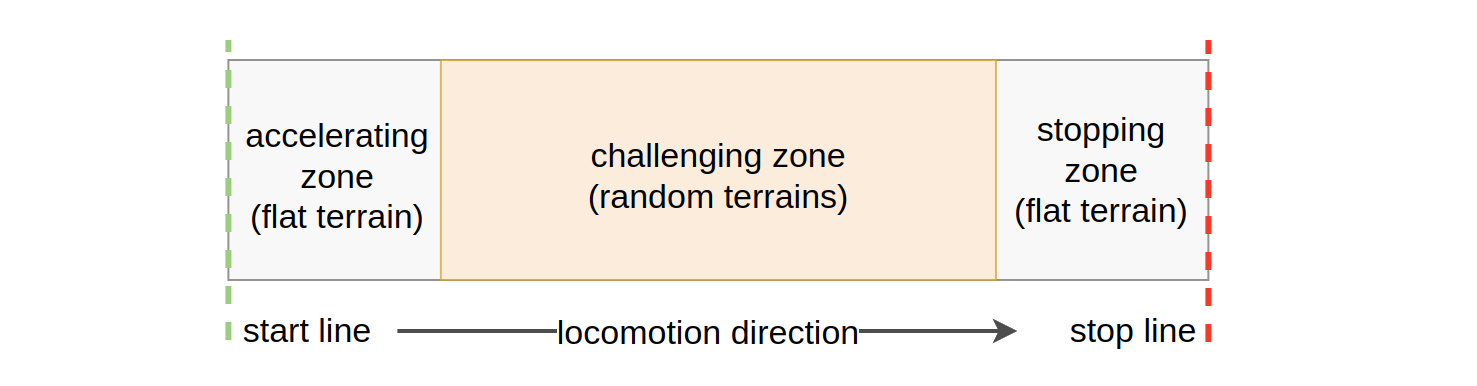}
    \caption{Training and testing environment for challenging terrains: simulated robot needs to accelerate to the target speed on a flat terrain, then go through a challenging zone with stairs and finally slow down to stop. The width of stairs are all equal to the width of the terrain and the maximum altitude change is $5$ cm for each stair.}
    \label{fig:testenv}
\vspace{-10px}
\end{figure}

We designed three simulated experiments \revised{(Figure \ref{fig:teaser})} to explore the answer to the research questions proposed. In each experiment, we evaluate the learned policies on the following tasks:
\begin{enumerate}
    \item \textbf{VEL}: \revised{Random velocity profiles ($v_{max} \in [0.5, 0.9]$ m/s)}
    \item \textbf{PER}: \revised{Random perturbations ($F_z \leq 20$N, $F_{xy} \leq  10 $N)}
    \item \textbf{STR}: \revised{Random stairs ($\Delta_{altitude} \leq 5$ cm, length $\leq 1$ m)}
\end{enumerate}
\revised{where $F_z$, $F_{xy}$ are vertical and horizontal external forces on the torso. $\Delta_{altitude}$ is the maximum altitude change between neighbor stairs.} The first two experiments share the same policies trained on the solely flat terrain, while policies for the last experiment are trained on challenging terrains described in Figure \ref{fig:testenv}. All experiments last for $25$ seconds. These three experiments are designed for three types of scenarios separately: a flat terrain without perturbations (\emph{Flat/Stationary}), a flat terrain with perturbations (\emph{Flat/Perturb}) and a random terrain without perturbations (\emph{Random/Stationary}). 

We evaluate three tasks based on the following criteria. For \textbf{VEL} test, we use average velocity error $\bar{E}_v$ to evaluate whether agents are capable of following different velocity profiles.  For \textbf{PER} and \textbf{STR} tests, we use average travel distance $\bar{D}$ to evaluate whether agents are able to keep stable locomotion under unexpected perturbations or obstacles. The expected travel distance was $15.0$ m for both tests.

\noindent \textbf{Results.}
First, we compared the learning curves in two different training terrains of four agents, PMTG, PMTG-contact, PM-FSM (ours), and PM-FSM-reflex (ours), in Figure \ref{fig:trainingcurve}. The results indicate that PM-FSM and PMTG-contact agents achieved much higher rewards than PMTG on the flat terrain by closely following the desired velocity profile. PM-FSM agents have even larger rewards ($\ge 0.58/0.6$) than PMTG-contact. Our PM-FSM-reflex is the best agent in both challenging tasks, \textbf{PER} and \textbf{STR}. For all stair-specific training, our PM-FSM-reflex is the only one that learns effective policies to go through the whole challenging scenarios with stairs and random perturbations.

We then evaluated the robustness of all the three agents by deploying them to  
\revised{scenarios that} are unseen during training. We conducted ten tests by randomly generating velocity profiles, external perturbations, and stair configurations.

\begin{table}[htbp]
 
\caption{Results of Simulated Test Experiments }
\begin{tabular}{c|c|c|c}
\hline
\textbf{Experiment} &\textbf{VEL} & \textbf{PER} & \textbf{STR} \\ 
\hline
\textbf{Scenarios} &  Flat/Stationary& Flat/Perturb & Random/Stationary \\
\hline
\vspace{-0.75em}
& & & \\
\textbf{Measurements}&\textbf{$\bar{E_v}$ $(m/s) \downarrow$}  & \textbf{$\bar{D}$ $(m) \uparrow$} &  \textbf{$\bar{D}$ $(m) \uparrow$}\\
\hline
PMTG\cite{iscen2019policies} & 0.254 & 7.38 & 0.84 \\
PMTG-contact & 0.115 & 9.20 &  1.25 \\
PM-FSM & \textbf{0.055} & 8.35 & 1.71 \\
PM-FSM-reflex & 0.064 & \textbf{10.25} & \textbf{5.34}\\
\hline 
\end{tabular}
\label{tab1}
\vspace{-5pt}
\end{table}

Table \ref{tab1} presented the comparison results \revised{obtained in the tested unseen scenarios}. For the \textbf{VEL} task, the results show both PM-FSM agents were able to produce velocities closely according to the input target velocities, while both PMTG and PMTG-contact agents resulted in much larger velocity errors. This trend indicates that PM-FSM performs more robust on unseen tasks because the training performances of PMTG-contact, PM-FSM, and PM-FSM-reflexes are comparable in Figure~\ref{fig:trainingcurve}.
For the \textbf{PER} test, PM-FSM-reflex agent performed the best to resist the impacts from random perturbations. For the \textbf{STR} scenario, PM-FSM with reflexes was the only agent that managed to go through most random stair terrains.

These results support our hypothesis that the explicit notion of contact flags (PM-FSM, PM-FSM-reflex) is helpful to obtain more robust agents compared to the zero awareness of contacts (PMTG) or the naive formulation (PMTG-contact). The results also suggest that the designed reflex mechanism helped the agent to overcome challenging terrains as well as obtain more robust policies.

\begin{figure}[!ht]
    \centering
    \includegraphics[width=0.48\textwidth]{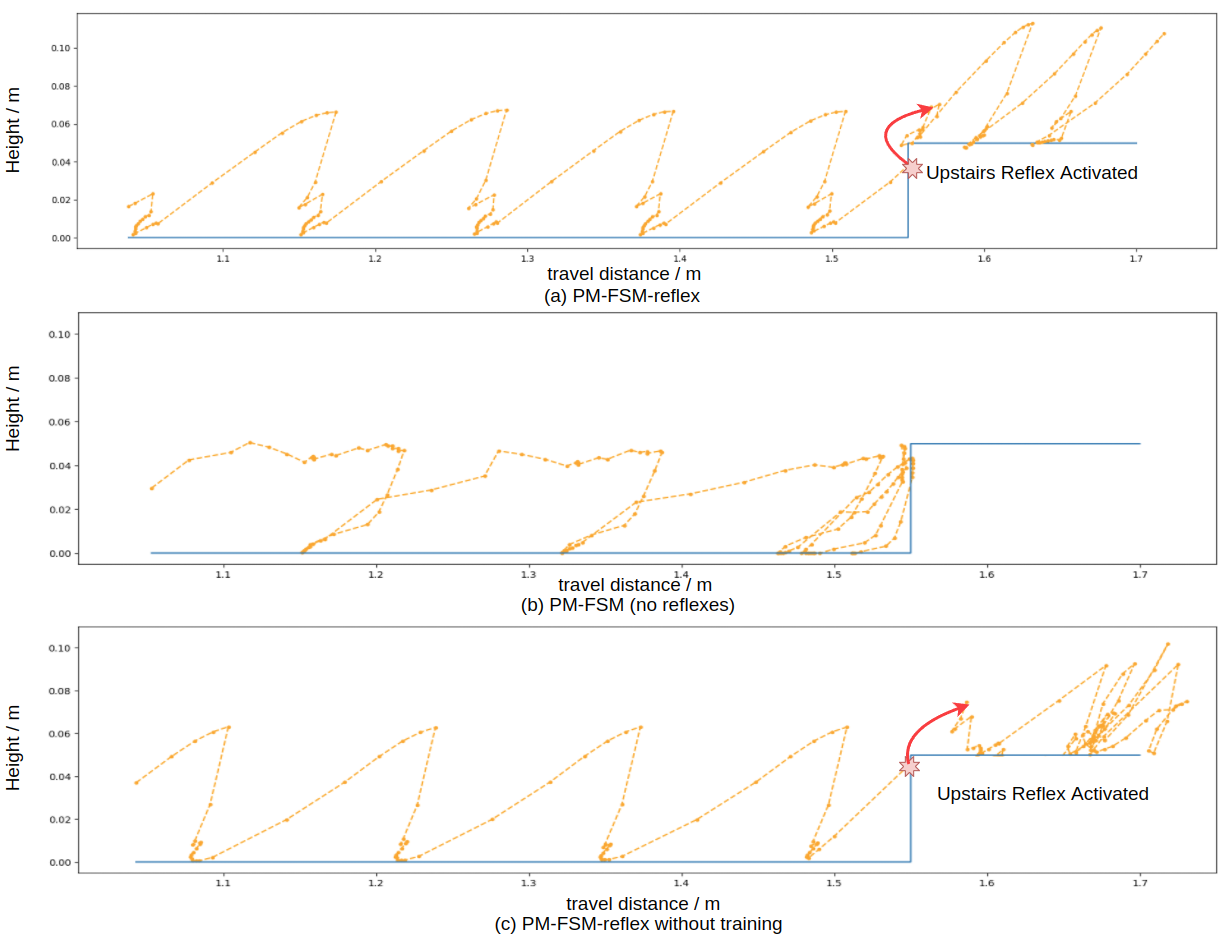}
    \caption{FR foot joint trajectories: (a) PM-FSM-reflex, (b) PM-FSM, (c) PM-FSM without training. We use points to track the foot joint positions every control interval, and use dashed lines to show approximate trajectories between control intervals.}
    \label{fig:learned-reflex}
    \vspace{-1em}
\end{figure}

For a more detailed analysis, we plotted the foot joint trajectory of the front right leg (FR) in both flat and challenging terrains to visualize how the PM-FSM-reflex agent managed to overcome stairs (Figure~\ref{fig:learned-reflex}). We also plotted the corresponding trajectories using an untrained PM-FSM-reflex and a trained non-reflex PM-FSM. We observed that non-reflex PM-FSM was unaware of heights and only learned to keep balance. Although the untrained PM-FSM-reflex successfully put the FR leg on the stair, it failed to lift the whole body onto the stairs. On the other hand, PM-FSM-reflex learned to put the foot joint a little backward to raise its torso and go through the stairs. This result supports the claim that our PM-FSM-reflex well learns to modulate the behaviors of the reflexes to overcome various stairs.

\subsection{Real Robot Experiments}

\begin{figure}[!h]
    \centering
    \includegraphics[width=0.45\textwidth]{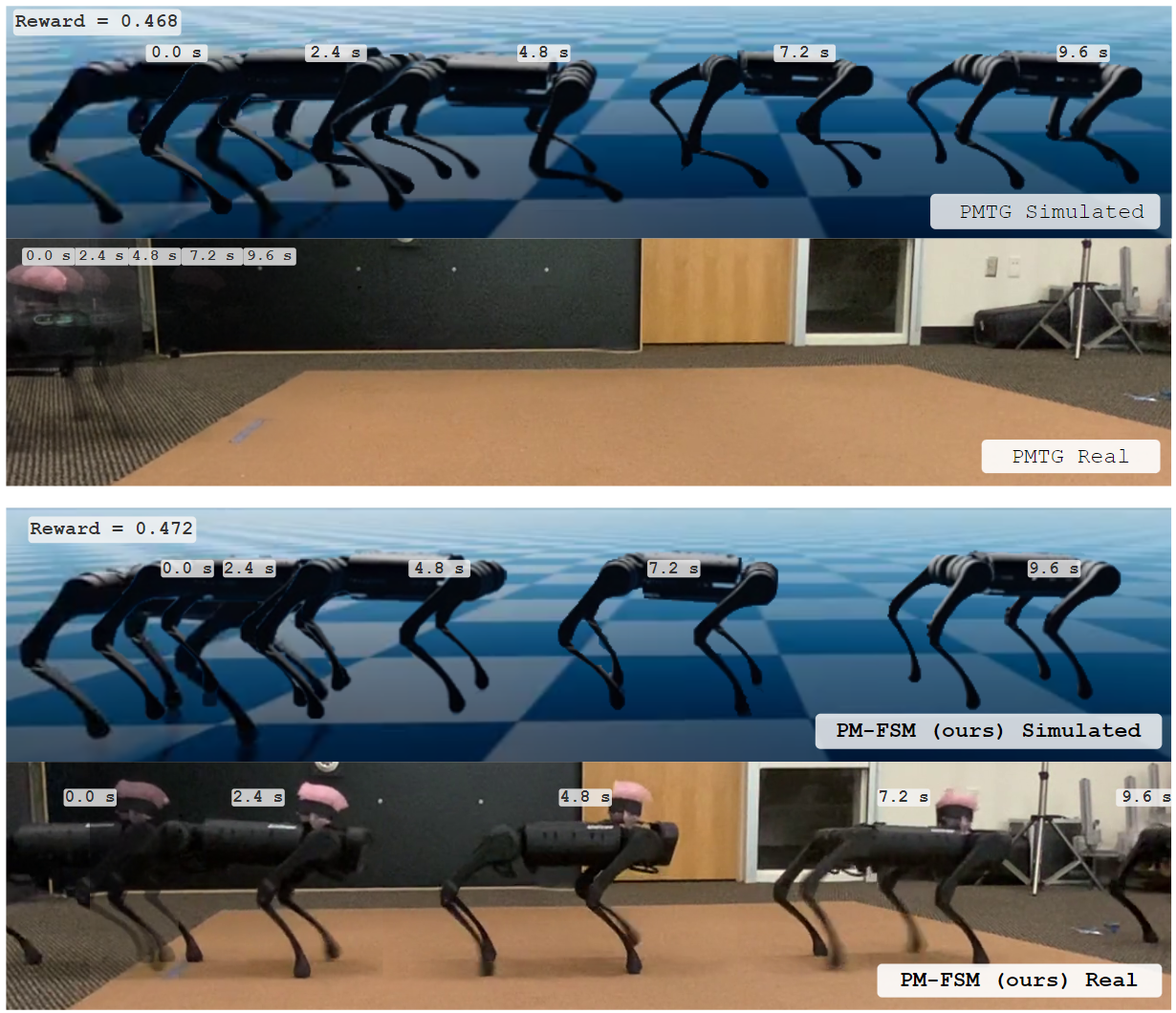}
    \caption{Result of simulated and real robot experiments. Time tickets at $0.0s$, $2.4s$, $4.8s$, $7.2s$, and $9.6s$ are chosen for recording the robot's positions.}
    \label{fig:real}
    \vspace{-10pt}
\end{figure}

We also deployed the policy learned in the simulated environments to the real A1 robot to test whether our approach is robust enough to handle the sim-to-real gap. The main challenges for the sim-to-real transfer are control latency and observation noise. To deal with the challenges, we import domain randomization in the training process by randomizing some "missing" control steps, randomizing the position of the center of mass, and randomizing the external perturbation. Both policies are trained under exactly the same conditions (number of total rollouts, random seed, etc.). In the real robot experiments, we set the test maximum velocity at $0.5$ m/s with a control frequency of $20$ Hz. Figure \ref{fig:real} shows the comparison between the real robot experiments and the corresponding simulated experiments. From the results, we observed that the original PMTG model failed to produce a successful forwarding policy. And our PM-FSM model moved forward to the target, for contact-aware FSM controller can be more robust against unpredictable mechanical latency. 

Please note that this experiment does not disprove the sim-to-real transferability of the original PMTG. If we use more complex training approaches or carefully tune DR parameters, it would be possible to obtain a reasonable PMTG policy that works on the hardware. However, we observed that our PM-FSM could better bridge the sim-to-real gap without careful domain randomization tuning.


\vspace{-0.2em}
\section{Conclusion}

\vspace{-0.2em}
In this work, we propose a novel policy architecture, Policies Modulating Finite State Machines (PM-FSM), that takes advantage of the prior knowledge of locomotion contexts and the associated walking automata. Our key idea is to extend the trajectory generator with a finite state machine, as its name suggests. This invention allows us to learn a robust locomotion policy by being explicitly aware of the desired event sequences. We evaluate the proposed method on various challenging locomotion tasks with randomized terrains and perturbations. We demonstrate that our method can outperform two baselines: PMTG and PMTG with contact flags, which demonstrates the importance of the explicit awareness of contacts. We also succeed in the sim-to-real transfer from the RaiSim environment to the A1 robot. 

Based on our current results, we plan to extend the current FSM design to learn more robust quadrupedal locomotion. In this work,  our contact-aware FSM is based on simple ideas proposed by Mcghee \cite{mcghee1968some}. In future work, we are looking forwards to importing advanced design of FSM-based locomotion controllers to our architecture as more powerful modules.
We believe this extension will give a policy the ability to sense a broad range of events more explicitly, eventually leading to better robust behaviors in many challenging environments.

\section*{Acknowledgements}
This work is supported by the National Science Foundation under Award \#2024768.
We would like to thank Visak C.V. Kumar for his help during the real world deployment.
\vspace{10px}
{
\balance
\small
\bibliographystyle{ieee_fullname}
\bibliography{fsm-pmtg}
}

\end{document}